\documentclass[conference]{IEEEtran}
\IEEEoverridecommandlockouts

\usepackage{cite}
\usepackage{amsmath,amssymb,amsfonts}
\usepackage{algpseudocode}
\usepackage{algorithm}
\usepackage{graphicx}
\usepackage{textcomp}
\usepackage{xcolor}
\def\BibTeX{{\rm B\kern-.05em{\sc i\kern-.025em b}\kern-.08em
    T\kern-.1667em\lower.7ex\hbox{E}\kern-.125emX}}
\begin{document}

\title{Regionalized Metric Framework: A Novel Approach for Evaluating Multimodal Multi-Objective Optimization Algorithms}

\author{\IEEEauthorblockN{1\textsuperscript{st} Jintai Chen}
\IEEEauthorblockA{\textit{dept. School of Engineering} \\
\textit{University of Western Australia}\\
Perth, Australia \\
23833705@student.uwa.edu.au}
\and
\IEEEauthorblockN{2\textsuperscript{nd} Fangqing Liu}
\IEEEauthorblockA{\textit{dept. School of Management} \\
\textit{Guangdong University of Technology}\\
GuangZhou, China \\
peerfog@gdut.edu.cn}
\and
\IEEEauthorblockN{3\textsuperscript{rd} Xueming Yan}
\IEEEauthorblockA{\textit{dept. School of Information Science and Technology} \\
\textit{Guangdong University of Foreign Studies}\\
GuangZhou, China \\
yanxm@gdufs.edu.cn}
\and
\IEEEauthorblockN{4\textsuperscript{th} Han Huang}
\IEEEauthorblockA{\textit{dept. School of Software Engineering} \\
\textit{South China University of Technology}\\
GuangZhou, China \\
hhan@scut.edu.cn}
}

\IEEEoverridecommandlockouts            
\IEEEpubid{\makebox[\columnwidth]{%
978-8-3315-3431-8/25/\$31.00~\copyright~2025 IEEE\hfill}%
\hspace{\columnsep}\makebox[\columnwidth]{}}

\maketitle

\IEEEpubidadjcol

\begin{abstract}
    
This study aims to optimize the evaluation metric of multimodal multi-objective optimization problems using a Regionalized Metric Framework, which provides a certain boost to research in this field. Existing evaluation metrics usually use the reference set as the evaluation basis, which inevitably leads to reference set dependence. To optimize this problem, this study proposes an evaluation metric based on a Regionalized Metric Framework. The algorithm divides the set of solutions to be evaluated into three regions, and evaluates each solution according to a unique scoring function for each region, which is combined to form the evaluation value of the solution set. To verify the feasibility of this method, a comparative experiment was conducted in this study. The results of the experiment are roughly the same as the trend of existing indicators, and at the same time, it can accurately judge the advantages and disadvantages of points equidistant from the reference set. Our method provides a new perspective for further research on evaluation metrics for multimodal multi-objective optimization algorithms.

\end{abstract}


\section{Introduction}
In recent years, with the increased attention to multimodal multi-objective optimization problems, research on solving such problems has gradually deepened. Typical applications \cite{ref1} include the backpack problem, navigation, and rocket engine optimization. For multimodal multi-objective optimization problems, the goal is not only to efficiently find global Pareto solution sets, but also to find a diverse solution set that provides decision makers with more choices. A diverse solution set usually contains multiple global or local optimal solutions that are evenly distributed in the decision space and the objective space.

This paper focuses on how to evaluate the superiority or inferiority of the solution sets obtained by multimodal multi-objective optimization algorithms using evaluation metrics. The two key performance indicators are convergence and diversity. Convergence measures the proximity of the solution set to the theoretical Pareto optimal frontier, while diversity assesses the evenness of distribution and the spread of the solutions in both decision space and objective space.

More specifically, this paper concentrates on the study of evaluation metrics for the use of reference sets. The Inverted Generational Distance (IGD) metric is commonly used to balance convergence and diversity by using reference sets. 
\begin{equation}
    \label{IGDeq}
      IGD(P, P^*) = \frac{\sum_{x \in P^*} \min_{y \in P} \text{dist}(x, y)}{|P^*|}
    \end{equation}
    
where $P$ is the set of solutions obtained by the algorithm and $P^*$ is the set of reference points, and $dist(x,y)$ denotes the reference set $P^*$ is the point in the reference set $x$ to the solution set $P$ is the Euclidean distance between the points in the reference set and the points in the solution set $y$ the Euclidean distance between them. For the results of $IGD$, the smaller the value of $IGD$, the better the performance of the algorithm. Of course, there are some possible problems for this evaluation metric, such as the objectivity problem, the inability to specifically distinguish between convergence and diversity, and the possibility of ignoring some of the inferior solutions. However, IGD has limitations, such as not clearly distinguishing between convergence and diversity, and possibly ignoring inferior solutions during the evaluation process.

To address these issues, this paper introduces a partition scoring evaluation mechanism. This strategy clusters the reference set into different regions based on the seeds, and evaluates convergence and diversity through partition scoring. By distinguishing between global and local optimal solutions, this method provides a more detailed assessment of the algorithm’s performance and helps identify the strengths and weaknesses of each solution set.

\section{Related Work \& Motivation}
Recent advancements in multimodal multi-objective optimization have led to the development of several algorithms, such as MMODE\_ICD \cite{ref5}, DN\_MMOES \cite{ref6}, MO\_Ring\_PSO\_SCD \cite{ref7}, MMOEAC/DC \cite{ref9}, and HREA \cite{ref10}. These algorithms frequently use IGD (IGDX and IGDF) as evaluation metrics for comparing performance which have been introduced in previous chapter.
In addition to IGD, Hypervolume (HV) and Pareto Set Proximity (PSP) are also considered. HV measures the volume dominated by the solution set in the objective space, giving an indication of the coverage of the Pareto frontier. PSP evaluates the proximity of the solution set to the true Pareto set, offering a more direct measure of convergence.

\begin{figure}[!t]
    \centering
    \includegraphics[width=3.5in]{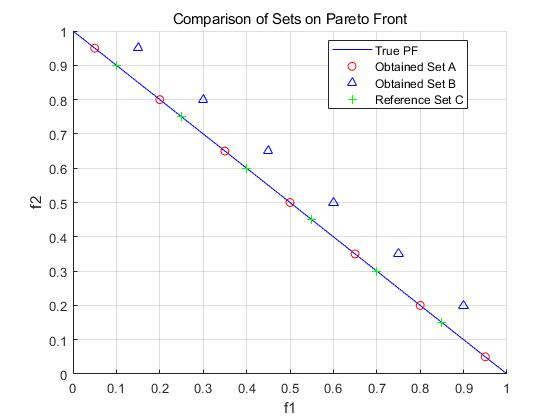}
    \caption{Objectivity Issues}
    \label{fig_21}
  \end{figure}
  
  From \cite{ref3} \cite{refhv} \cite{ref8}, it was found that commonly used evaluation metrics are often closely related to the Inverted Generational Distance (IGD). The advantages of the IGD metric (IGDX and IGDF) are as follows: First, it is comprehensive. IGD effectively evaluates both the diversity and convergence of the solution set derived from each algorithm, utilizing reference points. Second, it is simple and direct. Based on equation \eqref{IGDeq}, with a good reference set, IGD only requires calculating the minimum distance from each reference point to the solutions in the target set, and then computing the average of all these distances. This low computational cost allows for quick evaluation of the quality of the target solution set.

  However, there are several drawbacks to IGD. First, there is an issue with objectivity in the solution distribution. As shown in Fig. \ref{fig_21}, solution set A is closer to the true Pareto front than solution set B, and it should theoretically have a better evaluation. Yet, because both sets are equally distant from the reference set, their IGD scores are the same, leading to a lack of objectivity. 
  
\begin{figure}[!t]
    \centering
    \includegraphics[width=3.5in]{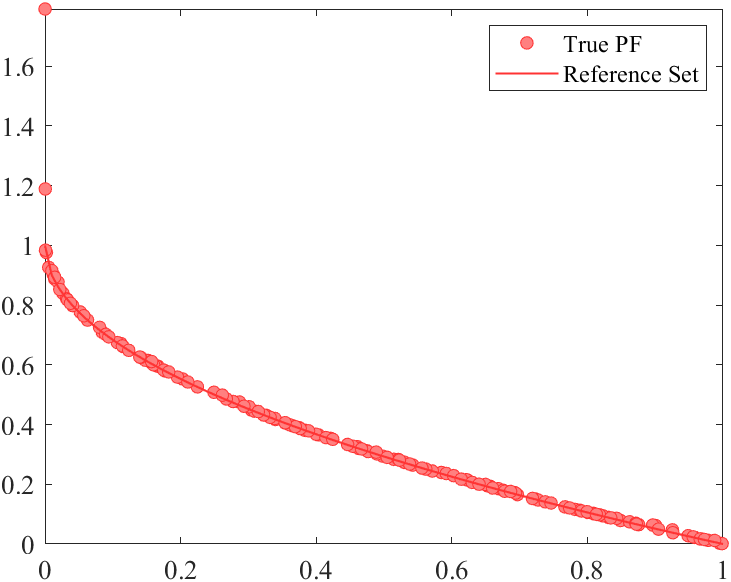}
    \caption{Visualization of MO\_Ring\_PSO\_SCD on MMF1 problem with target space solution set}
    \label{fig_22}
\end{figure}

  Second, IGD fails to distinguish between convergence and diversity performance. While IGD gives a comprehensive view of the algorithm's performance, it does not separately evaluate convergence and diversity. Finally, IGD may overlook inferior solutions. For example, as shown in Fig. \ref{fig_22}, the four inferior solutions on the vertical axis are too far from the reference points, and hence, IGD does not capture their distances. This issue can occur in any target solution set, causing IGD to neglect some of the less optimal solutions, further highlighting its lack of objectivity.
  
  Combining the above problems, this paper designed the evaluation metric based on Regionalized Metric Framework precisely from that perspective, which firstly determines the demarcation point of the concave and convex changes of each function through fine-tuning the reference set, and subsequently carries out clustering through the assistance of the points of the reference set, which is also an important operation of the region division, and then scores and integrates the solutions in each region, and finally output the convergence and diversity metrics of the target solution set. Finally, if researchers need to compare with other algorithms, the normalized weighted sum of the two metrics will be added for comparison; also, if researchers need to know the convergence and diversity of each optimal solution set, it will divide the clusters according to the pre-prepared sets of start and end points, and then compute the set of clusters divided by each set of clusters.

\section{Proposed Algorithm}
\subsection{Core formula for Regionalized Metric Framework}
In this subsection, this paper will analyze the formula of the region evaluation algorithm in detail, in the region evaluation algorithm, it will be divided into three evaluation regions - region 1 is given a score interval of $[2, 3]$; The scoring interval for region 2 is $[1,2]$ and the score interval of region 3 is $[0,1]$.

\subsubsection{Region 1 - Optimal Solution Region}
\begin{figure}[!t]
    \centering
    \includegraphics[width=3.5in]{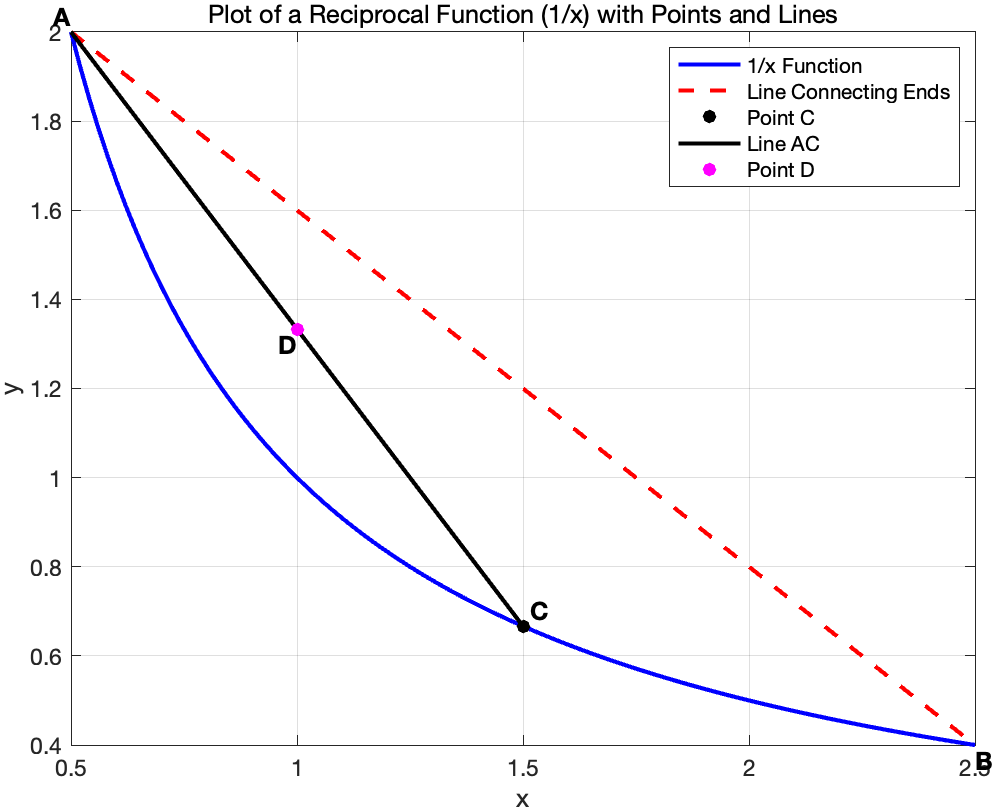}
    \caption{Example of a convex function optimal solution region}
    \label{fig_31}
  \end{figure}

First of all, for a reference point interval $[A,B]$, the convex function within a certain reference point interval, as shown in Fig. \ref{fig_31},
is recognized by the regional evaluation algorithm as the optimal solution region is the closed region composed of the segment function and the straight line AB.
For the points within the region, the scoring rules are as follows: as shown in Fig. \ref{fig_31}, 
assuming that the point D is a decomposition to be evaluated within the optimal solution region, first, the regional evaluation algorithm needs to approximate the length of the line segment AC that intersects the straight line AD and the convex function($d_1$)
\begin{equation}
  \label{d1}
  d_1 \approx \frac{\alpha}{\beta} \times d
\end{equation}
where $\alpha$ represents the angle between the tangent line at point A and the straight line AC, and $\beta$ represents the angle between the tangent line out of point A and the line AB, $d$ represents the length of the line segment AB. Subsequently, the process of calculating the score of the objective solution D is carried out.
The Regionalized Metric Framework uses the idea of approximate estimation in giving the score, which assumes that the midpoint of AC is the furthest point from the objective function on the line segment AC, so that the to-be-scored
solution D should be rated higher the farther it is from the midpoint.
In addition, the regional evaluation algorithm also combines the above points in this section
The mentioned $\alpha$ and $\beta$, because, if the $\alpha$ is larger, the farther the segment AC is from the target function, the score should also be lower, the opposite should be higher.Combining these two descriptions of the scoring rules, the scoring formula for the region of convex function superior solutions is:
\begin{equation}
  \label{region1}
  grade =
\begin{cases} 
    2 + \frac{2 \times d_2 - d_1}{d_1} \times \frac{\alpha}{\beta} & d_2 > \frac{d_1}{2} \\
    3 - \frac{2 \times d_2}{d_1} \times \frac{\alpha}{\beta} & d_2 \leq \frac{d_1}{2}
\end{cases}
\end{equation}
In this equation $\alpha$, $\beta$ and $d_1$ as shown above, the $d_2$ represents the distance from the target solution D to the reference point A.

\begin{figure}[!t]
    \centering
    \includegraphics[width=3.5in]{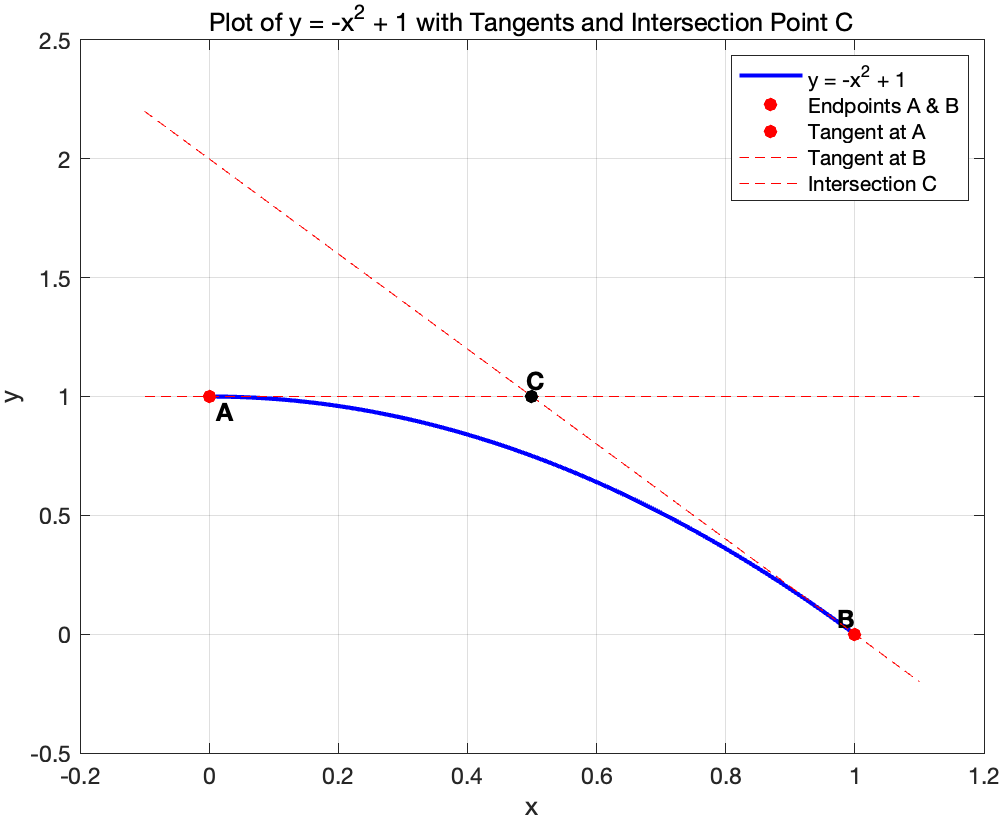}
    \caption{Example of a suboptimal solution region for a convex function}
    \label{fig_32}
  \end{figure}

For a concave function in a certain reference point interval $[A,B]$, as shown in Fig. \ref{fig_32}, the region of superior solution recognized by the region evaluation algorithm is the closed region formed by the segment of the function and its tangent lines at points A and B. The region of superior solution is the region of the tangent line to the function.
The scoring rules for this region are the same as those in Region 2. It should be noted that the final score should be added 1 based on the score calculated in Region 2. 
It should be noted that, because the region one algorithm needs to be combined with the derivative solution, considering the complexity of the derivation of the three-dimensional optimal function, at present, the region one algorithm is only applicable to the two-dimensional optimal function.

\subsubsection{Region 2 - Suboptimal Solution Region}
\begin{figure}[!t]
  \centering
  \includegraphics[width=3.5in]{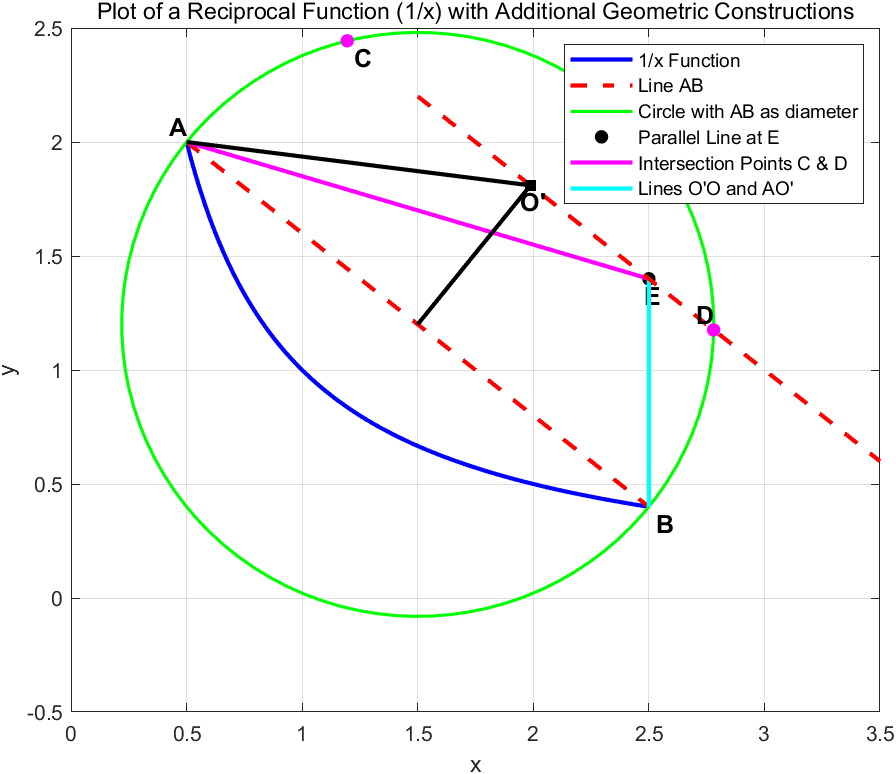}
  \caption{Example of a convex function optimal solution region}
  \label{fig_33}
\end{figure}

In the region evaluation algorithm, the giving mechanism of the two-dimensional and three-dimensional optimal function and the concave-convex function in the suboptimal solution region is the same, so in this chapter, the author takes the two-dimensional convex function as an example to introduce the giving mechanism of the suboptimal solution region.
As shown in Fig. \ref{fig_33},
in the reference set interval $[A,B]$, the circle drawn with AB as the diameter is the region where the target solution set is clustered, and the suboptimal solution region is both above the line segment AB and the closed region consisting of the clustered circle.
The scoring formula for this region is as follows: assuming that the solution E to be scored is in the suboptimal solution region, the scoring criterion in this region is that, on the line segment CD parallel to line segment AB where E is located, when the solution tends to be more and more in the middle of line segment CD, it means that the target solution E is farther away from the objective function, so the regional evaluation algorithm approximates that the middle point of line segment CD is the most far away from the objective function on CD.
O' is the farthest point from the objective function on CD, and if the objective solution E is closer to the point O', its score should be lower, in addition, the farther the line segment CD is from AB, the score should be reduced accordingly, so combined with the above scoring rules, the suboptimal solution region for Convex functions is given a score formula:
\begin{equation}
  \label{Region21}
  grade =
\begin{cases} 
    2 - \frac{d}{2R} - \frac{\frac{\pi}{2} - \alpha}{\pi - 2\beta} & d_1 < d_2 \\
    1.5 - \frac{d}{2R} + \frac{\beta - \alpha}{2\beta} & d_1 \geq d_2
\end{cases}
\end{equation}
And the Concave function(since in the concave function the closer to the point O' the closer to the target optimal function function, so the formula should be the opposite)
\begin{equation}
  \label{Region22}
  grade =
\begin{cases} 
    2 - \frac{d}{2R} - \frac{\frac{\pi}{2} - \alpha}{\pi - 2\beta} & d_1 \geq d_2 \\
    1.5 - \frac{d}{2R} + \frac{\beta - \alpha}{2\beta} & d_1 < d_2
\end{cases}
\end{equation}
where $\alpha$ represents the angle between the line AE and the line AB, the $\beta$ represents the angle between the line AO' and the line AB, $d$ represents the length between the line CD and the line AB, R represents the radius of the clustered circle, and $d_1$, $d_2$ represents the distance from point E to points A and B, respectively.
\subsubsection{Region 3 - Generalized solution regions}
The Generalized solution region is defined as the target solution set $P$ that consists of all the solutions in the target solution set that have not been clustered together The mechanism for assigning scores to the solutions in this region is to group all the solutions of the solution set ($P'$). The mechanism for assigning scores to the solutions in this region is to find their minimum distance from the points in $P'$ to a point in the reference set.
\begin{equation}
  \label{region3}
  \text{dist}(x, R) = \min_{y \in R} \text{dist}(x, y)
\end{equation} 
where $x$ represents the solution in $P'$. Then all distances are normalized and the processed data are their respective $grade$.

\subsection{Convergence and diversity score formula}
\subsubsection{Convergence scoring equation}
In the set G of scores on the target solution set output by the region evaluation algorithm, the convergence score formula is the summation of all solution scores:
\begin{equation}
  \label{convergence}
  \text{Convergence} = \sum_{i=1}^{n} G_i
\end{equation}
where $n$ represents the number of all solutions.
\subsubsection{Diversity scoring formula}
The evaluation of the diversity of optimization algorithms in the region assessment algorithm uses the idea of comparing the variance with a fixed value, based on the reference set, the algorithm in this paper creates a clusters,
then for a target solution set P with N solutions, the diversity of the optimal case is that in each cluster there are $\frac{N}{a}$ solutions,
and each solution is the highest score of 3. We calculate the average score of each cluster in that best case is 3. Therefore, this paper's algorithm takes 3 as a fixed value k. We only need to calculate the variance of the average score C of all the clusters in the target solution set P with the fixed value k, and then we can roughly determine the diversity score of the target solution set P, which is given by:
\begin{equation}
  \label{diversity}
  \text{Diversity} = \frac{\sum_{i=1}^{n} (C_n - k)^2}{||C||}
\end{equation}
where $n$ represents the number of clusters in the clustering.
\subsubsection{Algorithm Comprehensive Performance Scoring Equation}
Here, the diversity score and convergence score of each algorithm are used to be normalized to get respectively $S_1$ and $S_2$.
After that, the weights are summed up to get the final result with the weighting factors $\alpha$ and $\beta$(both defaulted to 0.5)can be changed according to the different degree of importance attached to diversity and convergence by the researchers, the specific formula is:
\begin{equation}
  \label{score}
  \text{Score} = \alpha \times S_1 + \beta \times S_2
\end{equation}

\subsection{Algorithmic Process}
The evaluation metric of multimodal multi-objective optimization algorithm based on Regionalized Metric Framework is created for multimodal multi-objective optimization problems in order to optimize the problem created by reference sets.

This algorithm is mainly used in the two-dimensional space, in the three-dimensional space has not been dealt in detail and the region 1 and region 2 are merged to regard as the same scoring area.

\begin{algorithm}
    \caption{Simplified Clustering Operation}
    \begin{algorithmic}[1]
    \Statex \textbf{Input:} $pop$: Target solution set; $Ref$: Reference set; $Point\_set$: Set of start and end points
    \Statex \textbf{Output:} $C_1, C_2, \dots, C_n$: Clusters; $pop\_not\_in$: Unclustered solutions
    
    \For{each reference point pair in $Ref$}
        \State Calculate the cluster center and radius: $Center \leftarrow \frac{Ref_i + Ref_{i+2}}{2}$, $R \leftarrow \frac{1}{2} \times dist(Ref_i, Ref_{i+2})$
        \For{each $pop_i$ in $pop$}
            \If{$dist(pop_i, Center) \leq R$}
                \State Assign $pop_i$ to the current cluster $C_n$
            \EndIf
        \EndFor
    \EndFor
    \State \textbf{Return} $C_1, C_2, \dots, C_n$, $pop\_not\_in$
    \end{algorithmic}
\end{algorithm}

The flowchart of the overall algorithm is: 1) the objective solution set solved by the multimodal multi-objective optimization algorithm is clustered by the created reference set as shown in Algorithm 1 
2) and the solutions in the clusters are scored by the region assessment algorithm which has beed discussed in detail in previous section A and shown in Algorithm 2, and then the diversity, convergence and comprehensive performance of the algorithms are evaluated according to the method which has been shown in Section B and convergence and diversity of the local solution sets are evaluated according to the needs by pre-set observation point; 3) Finally, comparison of the advantages and disadvantages of scores among the algorithms outputs a ranking of all the algorithms in terms of convergence, diversity and comprehensive performance. 

\begin{algorithm}
    \caption{Grading Framework}
    \begin{algorithmic}[1]
    \Statex \textbf{Input:} $C_1, C_2, \dots, C_n$: Clusters; $pop\_not\_in$: Unclustered solutions; $Func\_type(Dim)$: Function's dimension
    \Statex \textbf{Output:} $C_1, C_2, \dots, C_n$: Clusters with grades; 
    \Statex $pop\_not\_in$: Unclustered solutions with grades
    
    \If{$Func\_type.Dim = 2$}
        \For{each cluster $C_i$ in $C_1, \dots, C_n$}
            \For{each point $Cluster$ in $C_i$}
                \State Calculate euclidean distances($e$), slopes and
                \State  derivatives, between reference points($refs$)
                \If{$f^{'}_1 \leq k$}
                \Comment{Concave or convex function}
                    \For{each seed in $Cluster$}
                        \State Calculate $e$ between $ref$
                        \State Grade the seed by corresponding 
                        \Statex \hspace{25mm} Regionalized Metric Framework
                    \EndFor
                \Else
                    \For{each seed in $Cluster$}
                        \State Same as 8-10 lines
                    \EndFor
                \EndIf
            \EndFor
        \EndFor
    \Else
        \State Use only Regionalized Metric Framework in $Region ~2$ to 
        \Statex \hspace{4mm} grade solutions based on spatial vectors
    \EndIf
    \State For each solution in $pop\_not\_in$, find the nearest reference point and normalize the distance as the grade
    \State \textbf{Return} $C_1, C_2, \dots, C_n$ and $pop\_not\_in$
    \end{algorithmic}
\end{algorithm}
\subsection{Computational Complexity Analysis of RMF}

The time complexity of the proposed evaluation framework was analyzed based on two core algorithms (Algorithm 1 and Algorithm 2).
Let $n$ denote the number of solutions in the target set, $m$ denote the number of reference points, $d$ denote the dimension of the target space, and all single-step arithmetic operations be considered as $O(1)$.
Algorithm 1 clusters all reference point pairs by traversing them. In the worst case, the two-layer loop of Algorithm 1 requires $n$ distance comparison operations (dimension d) on $m-2$ reference point pairs (total number of solutions). Therefore, the total time complexity is: $O(m \cdot n \cdot d)$.
Algorithm 2 performs hierarchical clustering based on the clustering and nearest reference point pair solutions. Similar to the analysis of Algorithm 1, it involves two nested loops with the same number of sets. The worst-case complexity of the entire framework is also: $O(m \cdot n \cdot d)$. In summary, the complexity of this algorithm is $O(m \cdot n \cdot d)$.

\section{Experiments}
\subsection{Experimental setup}
This chapter will verify whether the evaluation metric of multimodal multi-objective optimization algorithm based on Regionalized Metric Framework can reasonably and objectively evaluate the multimodal multi-objective optimization algorithm. In this subsection, it will clarify the experimental environment and introduce the test set and its selection criteria.

\begin{enumerate}
    \item \textbf{Experimental Equipment:} All experiments were conducted on a desktop computer with a CPU of Intel(R) Core(TM) i7-14700KF at 3.40GHz and 64GB of running memory.
    
    \item \textbf{Experimental Software:} The experiments were run on PlatEMO developed with Matlab2020b as the development platform.
    
    \item \textbf{Algorithms:} The multimodal multi-objective optimization algorithms involved in the experiment are MMODE\_ICD and MMOEAC/DC. MMODE\_ICD are algorithms for convergent optimization, while MMOEAC/DC are algorithms for diversity optimization. The most significant difference between the convergent and diversity algorithms is that when solving a test problem with both locally optimal and globally optimal solutions, the convergent algorithms converge all the solutions to the Pareto first (globally optimal) level, while the diversity algorithms search both globally optimal and locally optimal solutions and eventually distribute the solutions evenly across the globally optimal and locally optimal solutions.
    
    \item \textbf{Test Problems:} From \cite{ref6} - \cite{ref13}, the study and the pre-experimental visualization observation of the test functions are screened based on the optimal functional form of the test functions, the convergence difficulty, and the existence of local and global optimal solution sets, and finally \cite{ref2} MMF1 (T1), MMF2 (T2), MMF4 (T3), MMF8 (T4), MMF9 (T5), MMF10 (T6), MMF13 (T7), and Omni-test (T8) as the representative test functions for this experiment.
\end{enumerate}

\subsection{Feasibility Analysis}
According to the data in the table \ref{table:convergence_scores} and \ref{table:diversity_scores}, the performance of different algorithms in terms of convergence and diversity is consistent with their focus. In terms of convergence, the MMODE\_ICD algorithm has excellent convergence in all test problems. 
MMOEAC/DC has a clear advantage in local diversity. In test problems such as T6 and T7 that contain local solutions, its diversity score shows a better even distribution, reflecting the effectiveness of the algorithm in dealing with problems with multiple local optima.
\begin{table}[ht]
    \centering
    \caption{MMODE\_ICD and MMOEAC/DC Convergence Scores}
    \renewcommand{\arraystretch}{1.5}
    \label{table:convergence_scores}
    \begin{tabular}{|c|c|c|}
    \hline
    \textbf{Test Problem} & \textbf{MMODE\_ICD} & \textbf{MMOEAC/DC} \\ \hline
    \textbf{T1} & \textbf{482.31 (2.40)} & 373.83 (13.98) \\ \hline
    \textbf{T2} & \textbf{309.57 (50.43)} & 184.85 (23.17) \\ \hline
    \textbf{T3} & \textbf{513.46 (3.47)} & 474.96 (28.06) \\ \hline
    \textbf{T4} & \textbf{441.53 (4.13)} & 378.61 (58.88) \\ \hline
    \textbf{T5} & \textbf{546.16 (5.92)} & 469.38 (4.00) \\ \hline
    \textbf{T6} & \textbf{563.26 (8.32)} & 478.30 (7.60) \\ \hline
    \textbf{T7} & \textbf{776.22 (10.25)} & 600.47 (57.93) \\ \hline
    \textbf{T8} & \textbf{711.84 (4.51)} & 156.00 (15.68) \\ \hline
    \end{tabular}
    \begin{quote}
      * Larger values in the convergence scores represent better convergence.
    \end{quote}
\end{table}

\begin{table}[ht]
    \centering
    \caption{MMODE\_ICD and MMOEAC/DC Diversity Scores}
    \renewcommand{\arraystretch}{1.5}
    \label{table:diversity_scores}
    \begin{tabular}{|c|c|c|}
    \hline
    \textbf{Test Problem} & \textbf{MMODE\_ICD} & \textbf{MMOEAC/DC} \\ \hline
    \textbf{T1} & \textbf{3.22 (0.19)} & 3.38 (0.29) \\ \hline
    \textbf{T2} & 6.85 (0.28) & \textbf{6.53 (0.39)} \\ \hline
    \textbf{T3} & \textbf{3.08 (0.23)} & 3.46 (0.38) \\ \hline
    \textbf{T4} & \textbf{5.52 (0.09)} & 5.73 (0.57) \\ \hline
    \textbf{T5} & 3.23 (0.13) & \textbf{3.13 (0.21)} \\ \hline
    \textbf{T6} & 6.16 (0.09) & \textbf{5.32 (0.27)} \\ \hline
    \textbf{T7} & 5.37 (0.16) & \textbf{4.94 (0.46)} \\ \hline
    \textbf{T8} & \textbf{4.20 (0.12)} & 8.16 (0.10) \\ \hline
    \end{tabular}
    \begin{quote}
      * Smaller values in the diversity scores represent better diversity.
    \end{quote}
\end{table}
Overall, the convergence and diversity scores in the table \ref{table:convergence_scores} and \ref{table:diversity_scores} match the focus of each algorithm. At the same time, the trend found in the calculation of the IGD values of the two algorithms, extracted from \cite{ref11}, is roughly the same as the experimental results in this paper. All of this shows that the evaluation matric based on the Regionalized Metric Framework is feasible.

\subsection{Analysis of Objective Optimization -- Case Study}
\begin{figure}[!t]
  \centering
  \includegraphics[width=3.5in]{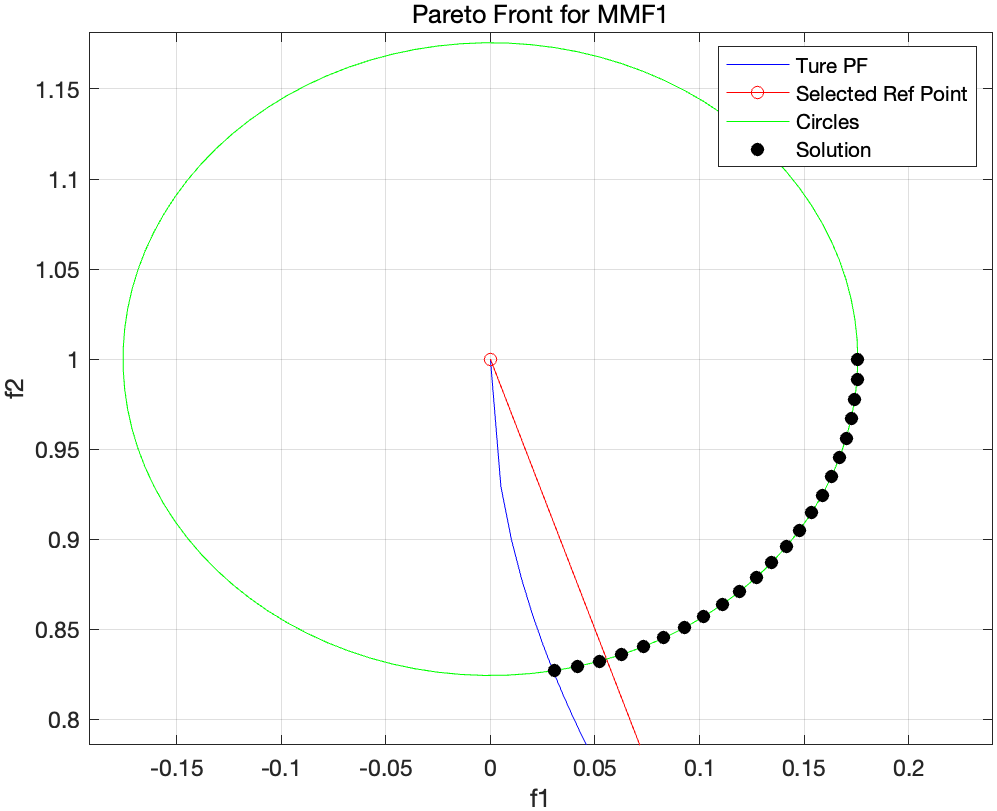}
  \caption{Example of equidistant points}
  \label{fig_45}
\end{figure}

In this section, by selecting six points that are the same distance from the reference point (0, 1), as shown in Figure \ref{fig_45}, the six black dots from left to right will be selected.
The aim is to analyze whether the algorithm in this paper can solve the possible limitations of the evaluation metric based on the reference set, especially when multiple solutions have the same distance from the reference point, and traditional evaluation methods have difficulty distinguishing between these solutions (e.g. IGD reflects that these solutions have the same evaluation value). 

From the data in the table \ref{table:specific_scores}, it can be observed that the scores of these points gradually decrease, with solution 1 scoring the highest (2.4576) and solution 6 scoring the lowest (1.4162), although their IGD scores were the same. This shows that although they have the same distance from the reference point, 
the evaluation index proposed in this paper gradually scores worse as the solution deviates from the objective function. 
This shows that the algorithm proposed in this paper has a certain degree of optimization in terms of objectivity compared to traditional reference set-based evaluation metrics.
\begin{table}[ht]
    \centering
    \caption{Specific scores for the selected 6 target solutions}
    \renewcommand{\arraystretch}{1.5}
    \label{table:specific_scores}
    \begin{tabular}{|c|c|c|c|c|}
    \hline
    \textbf{Spot Code} & \textbf{f1} & \textbf{f2} & \textbf{IGD} & \textbf{Score} \\ \hline
    1  & 0.0305 & 0.8270 & 0.0799 & 2.4576 \\ \hline
    2  & 0.0414 & 0.8293 & 0.0799 & 2.2603 \\ \hline
    3  & 0.0522 & 0.8322 & 0.0799 & 2.0630 \\ \hline
    4  & 0.0627 & 0.8359 & 0.0799 & 1.4784 \\ \hline
    5  & 0.0730 & 0.8402 & 0.0799 & 1.4470 \\ \hline
    6  & 0.0830 & 0.8451 & 0.0799 & 1.4162 \\ \hline
    \end{tabular}
    \end{table}
\section{Conclusion}
In conclusion, the paper has developed a multimodal multi-objective optimization algorithm evaluation metric based on a Regionalized Metric Framework, addressing the current research gap. The algorithm was fully implemented in Matlab and tested using the CEC2019 test set, validating the research hypothesis. The proposed evaluation index improves objectivity by overcoming the limitations of reference sets, allowing for the assessment of convergence and diversity in optimal solution sets, aiding future algorithm improvements.

Future work will focus on refining partition evaluation rules for higher-dimensional test problems and optimizing the use of derivatives, aiming to minimize them by adopting a more general region partitioning approach.

\vspace{12pt}
\end{document}